\documentclass[a4paper]{spie}  

\usepackage[english]{babel}
\usepackage{amsmath,amsfonts,amssymb}
\usepackage{graphicx}
\usepackage[colorlinks=true, allcolors=blue]{hyperref}
\usepackage{todonotes}
\usepackage{nicefrac}
\usepackage{bm}
\usepackage[caption=false]{subfig}
\usepackage{xcolor}
\usepackage{booktabs}
\usepackage[export]{adjustbox}
\usepackage{array}
\usepackage{multirow}
\usepackage[separate-uncertainty=true,detect-weight=true,detect-family=true]{siunitx}
\newcolumntype{N}{>{\centering\arraybackslash}m{2.75cm}}
\newcolumntype{M}{>{\centering\arraybackslash}m{2.25cm}}

\newcommand{\diag}{\mathop{\mathrm{diag}}}

\usepackage{tikz}
\usetikzlibrary{positioning,arrows}

\tikzset{
  >=stealth',
}

\title{Deep learning based 2.5D flow field estimation for maximum intensity projections of 4D optical coherence tomography}

\author[]{Max-Heinrich Laves}
\author[]{Sontje Ihler}
\author[]{L\"uder A. Kahrs}
\author[]{Tobias Ortmaier}
\affil[]{Leibniz Universit\"at Hannover, Institute of Mechatronic Systems, Appelstr. 11A, 30167 Hannover, Germany}

\authorinfo{Further author information: (Send correspondence to M.H. Laves)\\M.H. Laves: E-mail: laves@imes.uni-hannover.de, Telephone: +49\,511\,762\,19617}

\begin{document}
\maketitle

\begin{abstract}
In microsurgery, lasers have emerged as precise tools for bone ablation.
A challenge is automatic control of laser bone ablation with 4D optical coherence tomography (OCT).
OCT as high resolution imaging modality provides volumetric images of tissue and foresees information of bone position and orientation (pose) as well as thickness.
However, existing approaches for OCT based laser ablation control rely on external tracking systems or invasively ablated artificial landmarks for tracking the pose of the OCT probe relative to the tissue.
This can be superseded by estimating the scene flow caused by relative movement between OCT-based laser ablation system and patient.

Therefore, this paper deals with 2.5D scene flow estimation of volumetric OCT images for application in laser ablation.
We present a semi-supervised convolutional neural network based tracking scheme for subsequent 3D OCT volumes and apply it to a realistic semi-synthetic data set of ex vivo human temporal bone specimen.
The scene flow is estimated in a two-stage approach.
In the first stage, 2D lateral scene flow is computed on census-transformed en-face arguments-of-maximum intensity projections.
Subsequent to this, the projections are warped by predicted lateral flow and 1D depth flow is estimated.
The neural network is trained semi-supervised by combining error to ground truth and the reconstruction error of warped images with assumptions of spatial flow smoothness.
Quantitative evaluation reveals a mean endpoint error of $ (4.7\pm{}3.5) $\,voxel or \SI{27.5 \pm 20.5}{\micro\meter} for scene flow estimation caused by simulated relative movement between the OCT probe and bone.
The scene flow estimation for 4D OCT enables its use for markerless tracking of mastoid bone structures for image guidance in general, and automated laser ablation control.
\end{abstract}


\section{Description of purpose}
\label{sec:purpose}

Current minimally invasive cochlear implantation (MICI) techniques involve removal of the temporal bone by drilling, which demands high accuracy, as the approach to the inner ear passes sensitive structures\cite{Bergmeier2017,Weber2017}.
However, drilling can cause unintended damage to healthy tissue due to frictional heat or application of mechanical stress to delicate structures.
A promising alternative to this is the use of surgical lasers for contactless bone ablation in MICI, especially for the final step of bone removal, i.e. cochleostomy or extending the round window niche\cite{Kahrs2008}.
In ex vivo experiments the laser cochleostomy was done under monitoring with optical coherence tomography (OCT), which not only allows lateral control of the laser ablation with micrometer accuracy, but also supplies residual bone thickness measurements in depth\cite{Zhang2014a}.
The OCT information is highly valuable as it allows the prevision of sub-surface (risk) structures, which can be used to safely deactivate the laser beforehand.
Damaging false tissue in cochlear implantation can cause loss of residual hearing ability or malfunction of the nervus facialis and corda tympani, which can lead to facial paralysis or loss of taste.

The spatial control of the laser requires intraoperative navigation and registration from OCT to ablation laser and OCT to patient\cite{Fuchs2015,Stopp2008}.
In addition to the use of external optical navigation, which makes the process even more complex, OCT itself can be used as high-precision navigation system for tracking relative motion between the laser target area and the joint system consisting of ablation laser and OCT (AL-OCT).
OCT as optical tracking system has already been addressed.
However, the existing approaches either need artificial markers or the use of error-prone hand-crafted features\cite{Gessert2018,Laves2017}.
Artificial landmarks were created with the laser into the bone structure for tracking purpose\cite{Zhang2014b}.
Besides the invasiveness of this method, fixed landmarks have the disadvantage to get out of sight or change morphologically during necessary bone removal.
Therefore, this paper addresses the problem of tracking relative movement between an OCT acquisition probe and mastoid bone structures.
We present a method for markerless 2.5D voxel displacement (scene flow) estimation on 4D OCT images based on convolutional neural networks (CNN).

\begin{figure}[tb]
  \centering
  \begin{minipage}[c][6cm][c]{0.4\textwidth}
    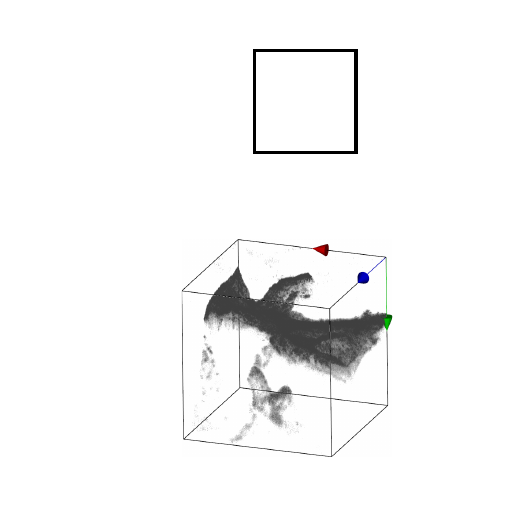
  \end{minipage}
  \quad
  \begin{minipage}[c][6cm][c]{0.4\textwidth}
    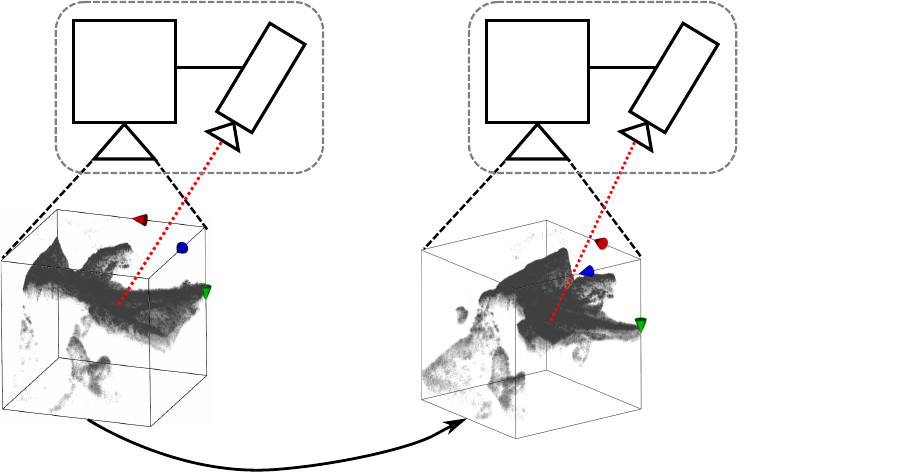
  \end{minipage}
  \caption{(a) Setup for ablation laser OCT setup. (b) The relative movement between AL-OCT system and tissue has to be considered for automated high accuracy laser bone ablation.}
  \label{fig:octlaser}
\end{figure}

\section{Methods}
\label{sec:methods}

\begin{figure}[tb]
  \centering
  \begin{minipage}[c][4.75cm][c]{1\textwidth}
    \centering
    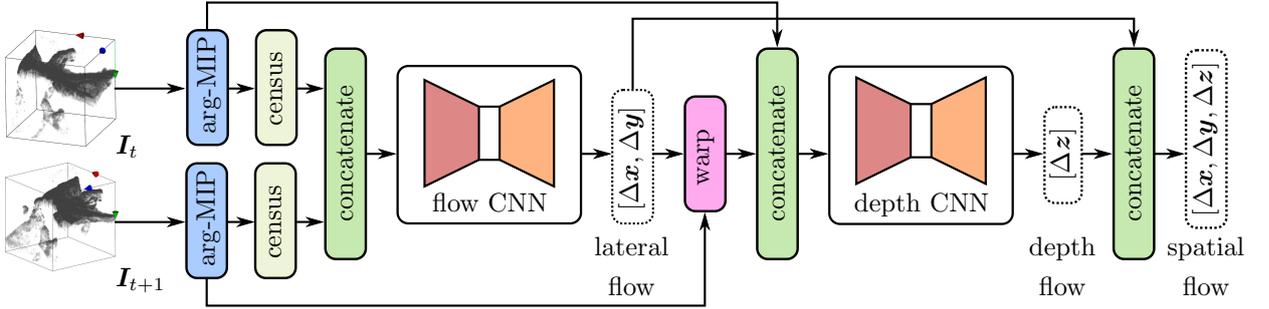
  \end{minipage}
  \caption{Basic structure of the proposed tracking scheme. Both the flow CNN and the depth CNN consist of a multi-stage pyramidal network as visualized in Fig.~\ref{fig:spynet}.}
  \label{fig:workflow}
\end{figure}

Convolutional neural networks have recently shown superior performance in processing OCT images, for example by identifying medical diagnoses on retinal OCTs with human performance\cite{Kermany2018} or in pose estimation with micrometer accuracy\cite{Gessert2018}.
CNNs have also lead to state-of-the-art performance in estimating 2D pixel displacements on subsequent camera images (optical flow)\cite{Ilg2017}.
Therefore, the approach proposed in this study extends CNNs for optical flow estimation and applies them to 4D OCT data for estimating voxel displacements over time.

An overview of the implemented workflow is shown in Fig.~\ref{fig:workflow}.
Two subsequent OCT volumes $ \bm{I}_{t} $ and $ \bm{I}_{t+1} $ are fed into separate branches of the tracking scheme.
In the first step, en-face depth maps using arguments-of-maximum intensity projections (arg-MIP) are calculated for each of the volumes by projecting the depth values of the maximum voxels along parallel rays onto image planes, yielding $ \bm{z}_{t} $ and $ \bm{z}_{t+1} $.
These depth maps can be interpreted as concise 2.5D representations of the original input volumes.
In order to compensate for depth changes between the two volumes, which will be considered by a second CNN later, the non-parametric census transform is applied to the depth maps, which results in $ \bm{z}^{c}_{t} $ and $ \bm{z}^{c}_{t+1} $.
Both subsequent representations $ \bm{z}^{c}_{t} $ and $ \bm{z}^{c}_{t+1} $ are then used for semi-supervised training a CNN $ F $, which estimates the dense planar optical flow between them.
The predicted 2D flow $ \left[ \Delta \bm{x}, \Delta \bm{y} \right] $ is used to warp $ \bm{z}_{t} $ onto $ \bm{z}_{t+1} $, resulting in $ \hat{\bm{z}}_{t+1} $.
After that, $ \hat{\bm{z}}_{t+1} $ and $ \bm{z}_{t+1} $ are fed into a second CNN $ D $ with similar architecture to $ F $, which estimates depth flow $ \left[ \Delta \bm{z} \right] $.
The results of $ F $ and $ D $ are concatenated to form the final output, which is referred to as 2.5D flow field $ \bm{V} = \left[ \Delta \bm{x}, \Delta \bm{y}, \Delta \bm{z} \right] $ with respect to the input volumes.

\subsection{Voxel displacement estimation}

\begin{figure}[tb]
  \centering
  \begin{minipage}[c][6.5cm][c]{1\textwidth}
    \centering
    \includegraphics[scale=1]{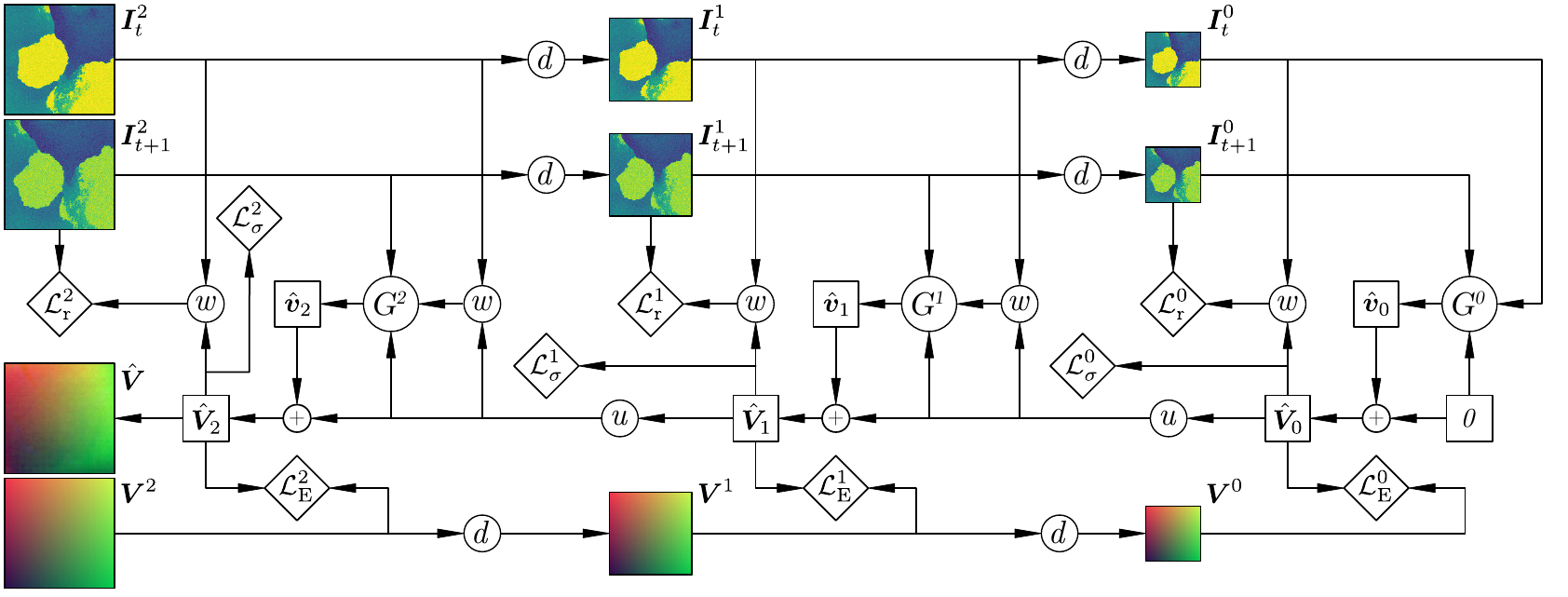}
  \end{minipage}
  \caption{Overview of a $ 3 $-stage pyramidal network for 2.5D optical flow estimation, which is inspired by SPyNet \cite{Ranjan2017}. The network $ G^{0} $ estimates initial flow using the images with lowest resolution. At the subsequent pyramid stages, $ G^{s} $ compute residual flow $ \bm{v}^{s} $ with regard to the upsampled previous one. After the last stage, the final flow $ \bm{V}^{S} $ is obtained. }
  \label{fig:spynet}
\end{figure}

The 2.5D flow field describes voxel displacements and changes in depth $ \bm{V} = \left[ \Delta \bm{x}, \Delta \bm{y}, \Delta \bm{z} \right] $ between two concise representations of successive OCT volumes, which is also referred to as \emph{scene flow} \cite{Huguet2007}.
In order to estimate the 2.5D flow field, a modified SPyNet architecture is used (see Fig.~\ref{fig:spynet}) \cite{Ranjan2017}.
The original SpyNet was designed to estimate optical flow between a consecutive pair of camera images and it comprises a coarse-to-fine pyramidal approach for estimating large motions, similar to the pyramidal Lucas-Kanade method\cite{Baker2004}.
It consists of multiple autoencoders $ G^{s} $, which estimate the residual optical flow $ \hat{\bm{v}}^{s} $ at different scales $ s $ of a scale pyramid of input images $ \bm{I}^{s}_{t} $ and $ \bm{I}^{s}_{t+1} $.
An image pyramid is built upfront by downsampling the input images to half size several times, where $ d $ denotes the downsampling process and $ u $ denotes the inverting (upsampling) operation.
The flow estimation at the highest stage $ s = 0 $ (smallest image) is used to warp $ \bm{I}^{0}_t $ onto $ \bm{I}^{0}_{t+1} $, yielding $ \hat{\bm{I}}^{0}_{t+1} $.
$ \hat{\bm{I}}^{0}_{t+1} $ is then upsampled to the next scale and, together with $ \bm{I}^{1}_{t+1} $, fed into $ G^{1} $, which estimates the residual flow on this stage.
The complete flow estimation
\begin{equation}
  \hat{\bm{V}}^{s} = \hat{\bm{v}}^{s} + u\left( \hat{\bm{V}}^{s-1} \right)
\end{equation}
is computed by combining the residual flow $ \bm{v}^{s} $ and the upsampled flow from the previous stage.
This is repeated for all scale stages until we obtain complete flow $ \bm{V}^{S} $ at full resolution.

We use the ErfNet autoencoder architecture for $ G^{s} $\cite{Romera2018}.
The network consists of a repeating basic module, which splits its input into two branches.
The main branch performs three convolutions with batch normalization and spatial dropout.
The outer convolutions reduce and increase the number of feature maps respectively, forming a ``bottleneck'' around the inner convolution.
Additionally, the convolutions are factorized into 1D convolutions.
The side branch just copies the input, which acts as a shortcut to the main branch.
The resulting layers drastically reduce computational cost and the number of parameters.
Similar to the connections in the ResNet\cite{He2016}, information can pass through the long-range connections, which avoids forming a data bottleneck.
Training of the flow field estimation network is done for every scale stage independently by utilizing a semi-supervised loss function $ \mathcal{L} $ described below.
All stages of the pyramid are trained individually until convergence.

\subsection{Census transform for depth invariance}

In order to compensate for depth differences between two input arg-MIPs of flow CNN $ F $, the non-parametric census transform is employed, which is briefly revisited in the following\cite{Demetz2015}.
The census transform is a kernel-based intensity order descriptor and is invariant to global intensity changes (which arg-MIPs correspond to at depth changes).
Every pixel of an image is represented by a binary vector, which encodes the spatial relationship of a center pixel with respect to the gray values of pixels in a local neighborhood.
Considering the 8 neighbors of a reference pixel in a $ 3 \times 3 $ neighborhood, every bit of the 8 bit binary vector encodes, which pixel has higher or lower intensity.
The census transform allows for separable lateral and distal voxel flow estimation as described above.
We implemented the census transform as layer operation for the machine learning library PyTorch 1.0 and published it as open source code.\footnote{\href{https://github.com/mlaves/census-transform-pytorch}{github.com/mlaves/census-transform-pytorch}}

%
%

\subsection{Semi-supervised loss function}

The proposed method is trained using both supervised and unsupervised criterions.
First, the ground truth flow $ \bm{v} $ from the semi-synthetic dataset is used and the endpoint error (EPE) loss $ \mathcal{L}_{\mathrm{E}}(\hat{\bm{v}}, \bm{v}) $ is calculated.
Ground truth flow for higher stages (smaller images) is calculated by downsampling the ground truth flow from full resolution.
Second, a reconstruction loss $ \mathcal{L}_{\mathrm{r}}(\hat{\bm{z}}_{t+1}, \bm{z}_{t+1}) $ is employed to maximize similarity between the second arg-MIP $ \bm{z}_{t+1} $ and the first arg-MIP $ \hat{\bm{z}}_{t+1} $ warped onto the second one.
This encourages the reconstructed image to have similar appearance to the corresponding input image.
Third, a smoothness loss $ \mathcal{L}_{\sigma} $ is used to encourage estimation of smooth flow\cite{Godard2017}.
The final loss function used to train the proposed method is
\begin{equation}
  \mathcal{L}^{s} = \alpha \mathcal{L}_{\mathrm{E}}^{s} + \beta \mathcal{L}_{\mathrm{r}}^{s} + \frac{\gamma}{2^{s-1}}\mathcal{L}_{\sigma}^{s} ~ .
\end{equation}
Loss $ \mathcal{L}^{s} $ is computed at each scale $ s $, resulting in the final training loss $ \mathcal{L} = \sum_{s=0}^{S-1} \mathcal{L}^{s} $.
The loss weights were empirically set to $ \alpha = 1.0 $, $ \beta = 0.5 $ and $ \gamma = 0.5 $.

\subsection{Semi-synthetic dataset}
\label{sec:dataset}

\begin{figure}
  \centering
  \subfloat[temporal bone specimen]{\includegraphics[height=3.8cm]{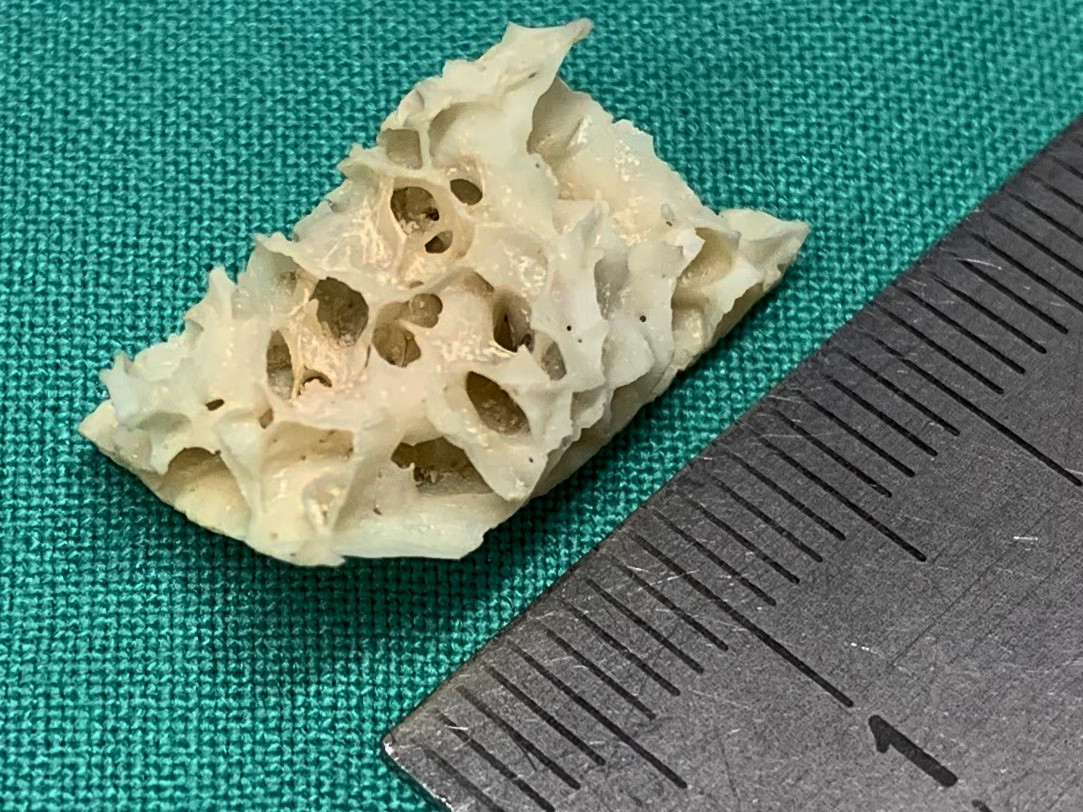}} \hfill
  \subfloat[OCT acquisitions]{ \includegraphics[height=3.8cm]{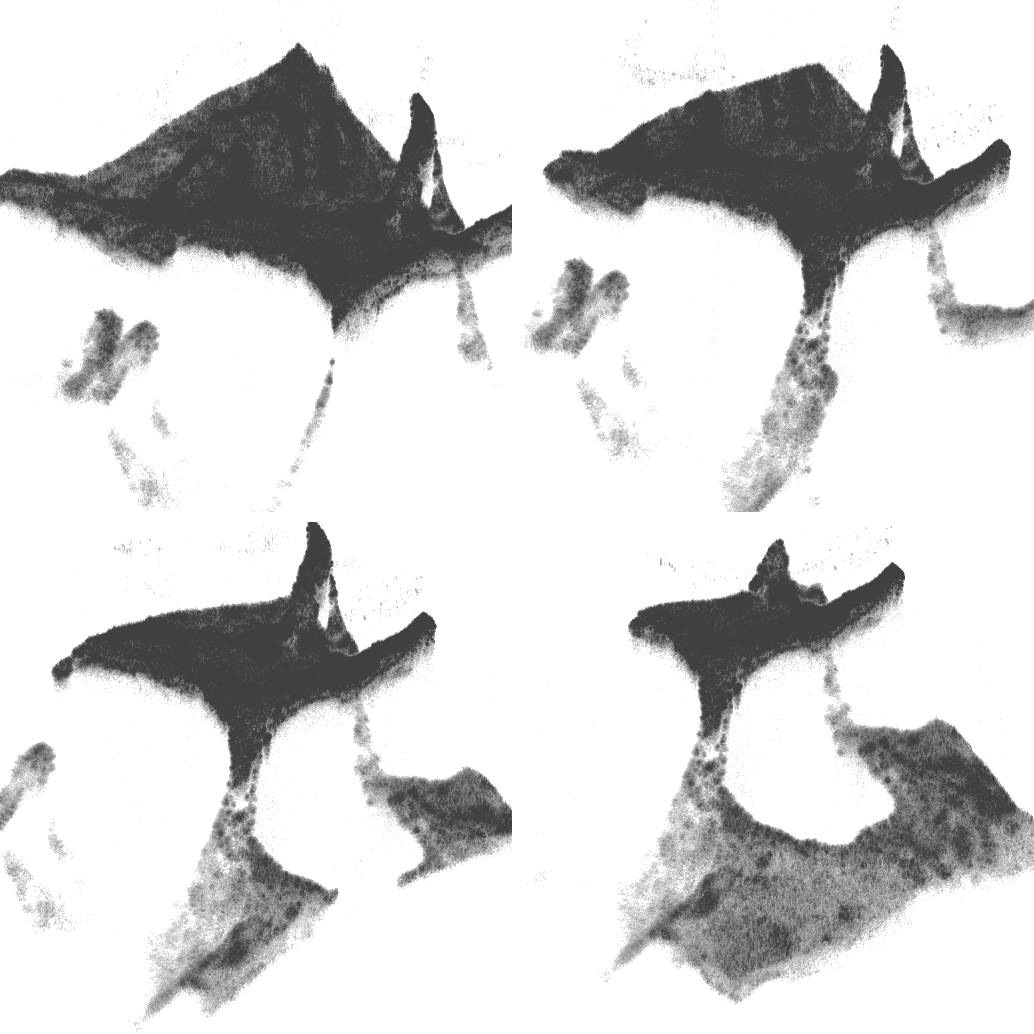}} \hfill
  \subfloat[en-face arg-MIP $ \bm{z} $]{\includegraphics[height=3.8cm]{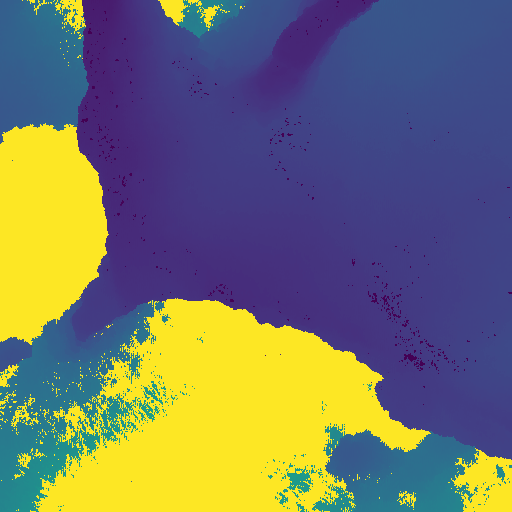}} \hfill
  \def\svgwidth{3.8cm}
  \subfloat[generated flow]{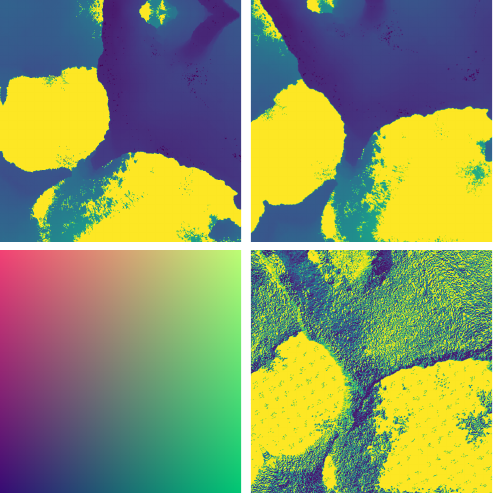}
  \caption{(a) Ex vivo human temporal bone with millimeter scale used for data set generation. (b) Volume rendering of different unique OCT acquisitions of ex vivo temporal bone. (c) En-face arg-MIP of an example OCT. (d) Example of adjacent image pair from generated ground truth flow and related census transformed image $ \bm{z}^{c} $. The normalized values of the 2.5D flow vectors are visualized as RGB image.}
  \label{fig:dataset}
\end{figure}

A data set of four unique 3D OCT acquisitions with different poses were created by placing ex vivo human temporal bone specimen (see Fig.~\ref{fig:dataset}\,(a--b)) on a 6-axis hexapod robotic platform (H824, Physik Instrumente GmbH \& Co. KG, Karlsruhe, Germany).
The OCT volumes were acquired with a swept-source OCT (OCS1300SS, ThorLabs Inc., lateral resolution \SI{12}{\micro\meter}).
The scan dimensions are equally set to 3\,mm for each spatial direction with a resolution of
512\,voxels.
This results in isotropic voxels with an edge size of \SI{6}{\micro\meter}.
Ground truth annotations for medical image data are difficult to obtain and are still an open problem\cite{Dosovitskiy2015}, especially for volumetric optical flow.
Therefore, we address this problem by generating warped image pairs with ground truth 2.5D displacements, which is common practice in 2D optical flow estimation, where various synthetic data sets exist\cite{Butler2012}.

Random affine transformations were applied to the arg-MIPs (Fig.~\ref{fig:dataset}\,(c)) of each of the OCT acquisitions with translations $ \bm{t} \sim \mathcal{N} \left( \bm{0}, \diag(\delta) \right) $ with $ \delta = 0.15 \cdot 512 $\,voxel along all spatial axes and rotations $ \omega \sim \mathcal{N} \left( 0, 0.25 \right) $\,radians around the vertical axis (see Fig.~\ref{fig:dataset}\,(d)).
This simulates relative movement between the OCT probe and the specimen.
Additionally, Gaussian noise were added to the pixel intensities of the warped arg-MIPs with $ \mu = 0 $ and $ \sigma = 0.1 $.
Each unique OCT is augmented to 1024 pairs of input images, which results in a total of 4096 image pairs.
We used 2048 image pairs from the first two OCT acquisitions as training set, and 1024 image pairs each from the third and fourth OCT for validation and testing, respectively.
In doing so, the validation set and the test set each contain images from OCT acquisitions that are not included in the training set.

\section{Results}
\label{sec:results}

\begin{figure}
  \centering
  \includegraphics{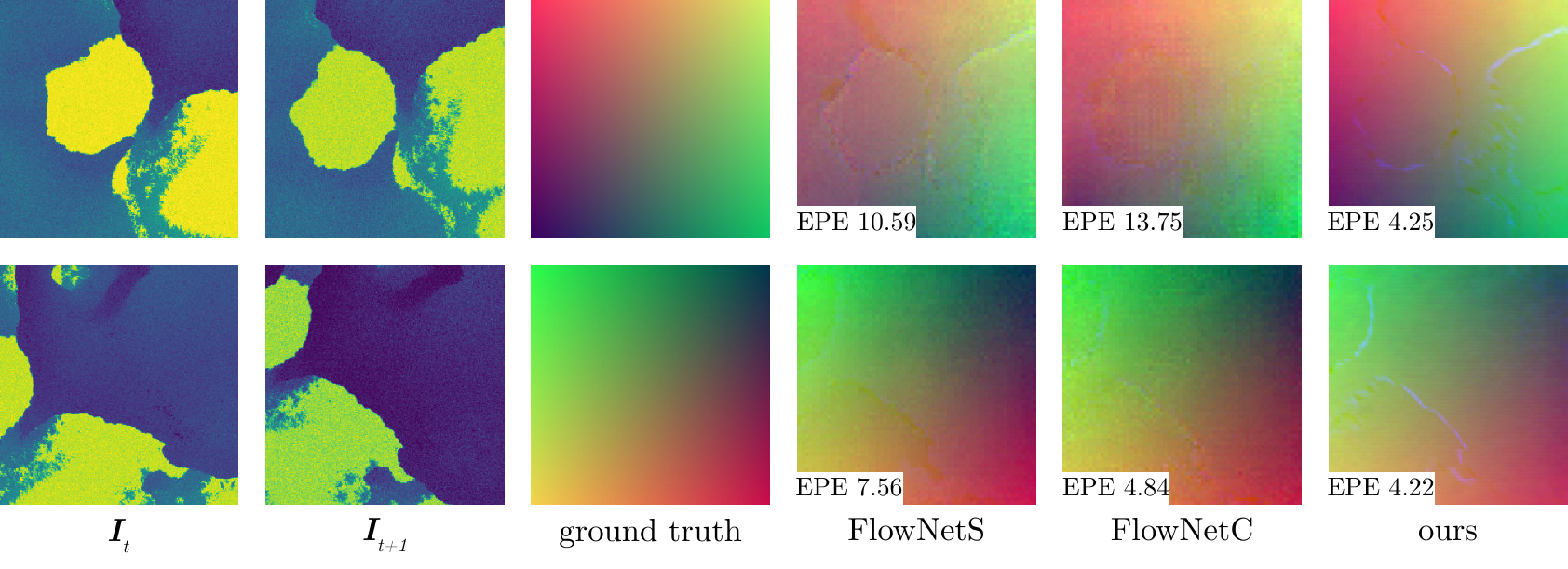}
  \caption{Qualitative results of our method compared to FlowNetS and FlowNetC.}
  \label{fig:results}
\end{figure}

\begin{table}[b]
  \centering
  \small
	\begin{tabular}{l N N N M}
          \toprule
          & \multicolumn{3}{c }{$ \mathcal{L}_{\mathrm{EPE}} $} & \multirow{ 2}{*}{\textbf{time} [ms]} \\
          \cmidrule(lr){2-4}
          & \textbf{training set} & \textbf{validation set} & \textbf{test set} &  \\
          \cmidrule(lr){2-5}
          FlowNetS       & \num{3.9 \pm 1.0} (3.7) & \num{5.4 \pm 1.9} (4.9) & \num{8.0 \pm 3.7} (6.9) & \textbf{8.4} \\
          FlowNetC       & \num{5.2 \pm 1.4} (5.1) & \num{6.4 \pm 1.9} (6.1) & \num{7.5 \pm 3.2} (6.5) & 10.8 \\
	  Ours & \textbf{\num{3.3 \pm 2.5} (2.7)} & \textbf{\num{3.5 \pm 2.0} (3.0)} & \textbf{\num{4.7\pm3.5} (3.8)} & 17.8 \\
	  Ours (unsupervised) & \num{3.6 \pm 1.8} (3.0) & \num{4.1 \pm 1.9} (3.5) & \num{6.0 \pm 4.2} (4.7) & 17.8 \\
	  \bottomrule \\
	\end{tabular}
	\caption{Mean (median) endpoint error in voxel compared to FlowNetS and FlowNetC\cite{Dosovitskiy2015}. Inference times were measured on a GeForce GTX\,1080\,Ti (Nvidia Corp., Santa Clara, CA, USA). Bold values denote best results.}
	\label{tab:results}
      \end{table}

Quantitative results are shown in Tab.~\ref{tab:results}.
As accuracy metric, we use the popularly accepted endpoint error (EPE) between predicted and ground truth flow.
To generate the results, we trained the presented approach semi-supervised with $ S = 4 $ scale stages on the training set for 200 epochs.
The weight configuration with the lowest loss value on the validation set was chosen (early stopping).
In order to compare to state-of-the-art methods, we trained FlowNetS and FlowNetC in the exact same manner\cite{Dosovitskiy2015}.
Our approach yields best results with a mean EPE of 4.7\,voxel on the test set, compared to FlowNetS with 8.0\,voxel and FlowNetC with 7.5\,voxel.
Figure~\ref{fig:results} shows qualitative results for two sample pairs.
All methods are able to correctly estimate 2.5D flow and show minor miscalculations at object boundaries and depth discontinuities.
FlowNetS has the lowest inference time of \SI{8.4}{\milli\second}.
With an inference time of \SI{17.8}{\milli\second}, real-time processing of data from current high-speed OCT devices with rates of more than 25\,volumes/s is also feasible with our approach.
Additionally, with a model size of 33.5\,MiB (8,252,664 parameters) and memory footprint during inference of 707\,MiB, compared to FlowNetS with model size of 156.8\,MiB (39,172,880 parameters) and memory footprint of 1207\,MiB, our method is more likely to be deployable to embedded medical devices.

\section{Conclusion}
\label{sec:conclusion}

Dense 4D scene flow is important for utilizing 4D OCT as an intra-operative modality for image-guided procedures such as laser bone ablation in minimally invasive cochlear implantation.
Current state-of-the-art high-speed OCT devices operate at 25 volumes per second \cite{Wieser2014} and the computational complexity of directly estimating volumetric scene flow were limiting its potential \cite{Laves2018}.

In this paper, we presented an approach for estimating 2.5D scene flow of 4D OCT by using concise arg-MIP representations.
The lateral and distal components of the scene flow are calculated subsequently by separating lateral and distal motion using the census transform.
Next, a deep CNN in a pyramidal structure is used to estimate residual scene flow in a coarse-to-fine approach.
By training the pyramid levels individually, the presented method provides superior results compared to FlowNetS and FlowNetC.
We achieved a mean endpoint error of 4.7\,voxel on the test set, which corresponds to an error of \SI{27.5}{\micro\meter}.
Our method therefore enables 3D OCT as a high-precision image-guidance tool that can operate faster than OCT acquisition rate (\SI{17.8}{\milli\second} per volume).
We envision, that the level-wise training of the pyramidal approach also makes unsupervised training feasible.
This is particularly important for training the method on real, non-synthetic data, since the generation of the ground truth in this case would require an error-prone hand-eye calibration.
In future work, laser experiments will be conducted on ex vivo specimens while the presented approach will be used to track and compensate relative motion between the laser-OCT system and the specimen.

\acknowledgments
This research has received funding from the European Union as being part of the EFRE OPhonLas project.

\bibliography{bibliography}

\begin{thebibliography}{10}

\bibitem{Bergmeier2017}
Bergmeier, J., Fitzpatrick, J.~M., Daentzer, D., Majdani, O., Ortmaier, T., and
  Kahrs, L.~A., ``{Workflow and simulation of image-to-physical registration of
  holes inside spongy bone},'' {\em {Int. J. Comput. Assist. Radiol.
  Surg.}}~{\bf 12}(8),  1425--1437 (2017).

\bibitem{Weber2017}
Weber, S., Gavaghan, K., Wimmer, W., Williamson, T., Gerber, N., Anso, J.,
  Bell, B., Feldmann, A., Rathgeb, C., Matulic, M., Stebinger, M., Schneider,
  D., Mantokoudis, G., Scheidegger, O., Wagner, F., Kompis, M., and
  Caversaccio, M., ``Instrument flight to the inner ear,'' {\em {Sci.
  Robot.}}~{\bf 2}(4) (2017).

\bibitem{Kahrs2008}
Kahrs, L.~A., Raczkowsky, J., Werner, M., Knapp, F.~B., Mehrwald, M., Hering,
  P., Schipper, J., Klenzner, T., and Wörn, H., ``Visual servoing of a laser
  ablation based cochleostomy,'' (2008).

\bibitem{Zhang2014a}
Zhang, Y., Pfeiffer, T., Weller, M., Wieser, W., Huber, R., Raczkowsky, J.,
  Schipper, J., W{\"o}rn, H., and Klenzner, T., ``{Optical coherence tomography
  guided laser cochleostomy: Towards the accuracy on tens of micrometer
  scale},'' {\em {BioMed Res. Int.}}~{\bf 2014} (2014).

\bibitem{Fuchs2015}
Fuchs, A., Pengel, S., Bergmeier, J., Kahrs, L.~A., and Ortmaier, T., ``Fast
  and automatic depth control of iterative bone ablation based on optical
  coherence tomography data,'' (2015).

\bibitem{Stopp2008}
Stopp, S., Deppe, H., and Lueth, T., ``{A new concept for navigated laser
  surgery},'' {\em {Lasers Med. Sci.}}~{\bf 23}(3),  261--266 (2008).

\bibitem{Gessert2018}
Gessert, N., Schlüter, M., and Schlaefer, A., ``A deep learning approach for
  pose estimation from volumetric oct data,'' {\em Med. Image Anal.}~{\bf 46},
  162--179 (2018).

\bibitem{Laves2017}
Laves, M.-H., Schoob, A., Kahrs, L., Pfeiffer, T., Huber, R., and Ortmaier, T.,
  ``{Feature tracking for automated volume of interest stabilization on 4D-OCT
  images},'' in [{\em Proc. SPIE}{\nolinebreak\hspace{0.1em}]},   {\bf 10135},
  10135--10135--7 (2017).

\bibitem{Zhang2014b}
Zhang, Y. and Wörn, H., ``{Optical coherence tomography as highly accurate
  optical tracking system},'' in [{\em {Proc. IEEE/ASME
  AIM}}{\nolinebreak\hspace{0.1em}]},   1145--1150 (July 2014).

\bibitem{Kermany2018}
Kermany, D.~S., Goldbaum, M., Cai, W., Valentim, C.~C., Liang, H., Baxter,
  S.~L., McKeown, A., Yang, G., Wu, X., Yan, F., et~al., ``{Identifying medical
  diagnoses and treatable diseases by image-based deep learning},'' {\em
  Cell}~{\bf 172}(5),  1122--1131 (2018).

\bibitem{Ilg2017}
Ilg, E., Mayer, N., Saikia, T., Keuper, M., Dosovitskiy, A., and Brox, T.,
  ``{FlowNet 2.0: Evolution of Optical Flow Estimation with Deep Networks},''
  in [{\em {Proc. IEEE CVPR}}{\nolinebreak\hspace{0.1em}]},   1647--1655 (July
  2017).

\bibitem{Ranjan2017}
Ranjan, A. and Black, M.~J., ``{Optical Flow Estimation Using a Spatial Pyramid
  Network},'' in [{\em {Proc. IEEE CVPR}}{\nolinebreak\hspace{0.1em}]},
  2720--2729 (July 2017).

\bibitem{Huguet2007}
Huguet, F. and Devernay, F., ``{A Variational Method for Scene Flow Estimation
  from Stereo Sequences},'' in [{\em Proc. ICCV}{\nolinebreak\hspace{0.1em}]},
   1--7 (2007).

\bibitem{Baker2004}
Baker, S. and Matthews, I., ``{Lucas-Kanade 20 Years On: A Unifying
  Framework},'' {\em {Int. J. Comput. Vis.}}~{\bf 56}(3),  221--255 (2004).

\bibitem{Romera2018}
Romera, E., Alvarez, J.~M., Bergasa, L.~M., and Arroyo, R., ``{ERFNet:
  Efficient Residual Factorized ConvNet for Real-time Semantic Segmentation},''
  {\em {IEEE Trans. Intell. Transp. Syst.}}~{\bf 19}(1),  263--272 (2018).

\bibitem{He2016}
He, K., Zhang, X., Ren, S., and Sun, J., ``{Deep Residual Learning for Image
  Recognition},'' in [{\em {Proc. IEEE CVPR}}{\nolinebreak\hspace{0.1em}]},
  770--778 (2016).

\bibitem{Demetz2015}
Demetz, O., Hafner, D., and Weickert, J., ``{Morphologically Invariant Matching
  of Structures with the Complete Rank Transform},'' {\em {Int. J. Comput.
  Vis.}}~{\bf 113}(3),  220--232 (2015).

\bibitem{Godard2017}
Godard, C., Aodha, O.~M., and Brostow, G.~J., ``{Unsupervised Monocular Depth
  Estimation with Left-Right Consistency},'' in [{\em {Proc. IEEE
  CVPR}}{\nolinebreak\hspace{0.1em}]},   6602--6611 (2017).

\bibitem{Dosovitskiy2015}
Dosovitskiy, A., Fischer, P., Ilg, E., Häusser, P., Hazirbas, C., Golkov, V.,
  v.~d. Smagt, P., Cremers, D., and Brox, T., ``{FlowNet: Learning Optical Flow
  with Convolutional Networks},'' in [{\em Proc. IEEE
  ICCV}{\nolinebreak\hspace{0.1em}]},   2758--2766 (2015).

\bibitem{Butler2012}
Butler, D.~J., Wulff, J., Stanley, G.~B., and Black, M.~J., ``{A naturalistic
  open source movie for optical flow evaluation},'' in [{\em Proc.
  ECCV}{\nolinebreak\hspace{0.1em}]},  {\em Part IV, LNCS 7577},  611--625
  (2012).

\bibitem{Wieser2014}
Wieser, W., Draxinger, W., Klein, T., Karpf, S., Pfeiffer, T., and Huber, R.,
  ``{High definition live 3D-OCT in vivo: design and evaluation of a 4D OCT
  engine with 1 GVoxel/s},'' {\em {Biomed. Opt. Express}}~{\bf 5}(9),
  2963--2977 (2014).

\bibitem{Laves2018}
Laves, M.-H., Kahrs, L.~A., and Ortmaier, T., ``{Volumetric 3D stitching of
  optical coherence tomography volumes},'' in [{\em {Proc.
  BMT}}{\nolinebreak\hspace{0.1em}]},   327--330 (2018).

\end{thebibliography}
\bibliographystyle{spiebib} 

\end{document}